\documentclass[runningheads]{llncs}

\usepackage[misc]{ifsym}
\usepackage[caption=false,font=footnotesize]{subfig}
\usepackage{amsmath}
\usepackage{cite}
\usepackage{amsmath,amssymb,amsfonts}
\usepackage{algorithmic}
\usepackage{graphicx}
\usepackage{textcomp}
\usepackage{comment}
\usepackage{multirow}
\usepackage{xcolor}
\def\BibTeX{{\rm B\kern-.05em{\sc i\kern-.025em b}\kern-.08em
    T\kern-.1667em\lower.7ex\hbox{E}\kern-.125emX}}

\begin{document}

\title{Cauchy Loss Function: Robustness Under Gaussian and Cauchy Noise\thanks{Supported by the NRF Thuthuka Grant Number 13819413.}} 

\author{Thamsanqa Mlotshwa \and
Heinrich van Deventer\orcidID{0000-0001-9309-4330} \and
Anna Sergeevna Bosman$^{\textrm{\Letter}}$\orcidID{0000-0003-3546-1467}}
\authorrunning{T. Mlotshwa et al.}
%
\institute{Department of Computer Science, University of Pretoria, South Africa\\
\email{thami.mlotshwa@gmail.com, hpdeventer@gmail.com, anna.bosman@up.ac.za}\\
}

\maketitle

\begin{abstract}
In supervised machine learning, the choice of loss function implicitly assumes a particular noise distribution over the data. For example, the frequently used mean squared error (MSE) loss assumes a Gaussian noise distribution. The choice of loss function during training and testing affects the performance of artificial neural networks (ANNs). It is known that MSE may yield substandard performance in the presence of outliers. The Cauchy loss function (CLF) assumes a Cauchy noise distribution, and is therefore potentially better suited for data with outliers. This papers aims to determine the extent of robustness and generalisability of the CLF as compared to MSE. CLF and MSE are assessed on a few handcrafted regression problems, and a real-world regression problem with artificially simulated outliers, in the context of ANN training. CLF yielded results that were either comparable to or better than the results yielded by MSE, with a few notable exceptions.

\end{abstract}

\begin{keywords}
mean squared error (MSE), Cauchy loss function (CLF), loss functions, generalisation, outliers
\end{keywords}

\section{Introduction}
Use of the mean squared error (MSE) loss function in the context of regression problems has been and remains fairly common owing to convention and its computational simplicity~\cite{Zahra2014}. MSE depends on the assumption that noise in the data has a Gaussian distribution \cite{Heravi2018}. But in practise, it is unlikely that the true noise distribution can be determined without a thorough understanding of the underlying data generating process~\cite{El-Melegy2009}. 

MSE tends to perform poorly in the presence of residuals that deviate significantly in size from the trend of the data~\cite{El-Melegy2009}. Large residuals can exceed the expected bounds that the assumption of Gaussianity imposes. This is often seen in real world problems, where particularly large residuals or outliers account for 1\% to 10\% or even more of typical datasets~\cite{El-Melegy2009}. There exist a number of mechanisms for handling outlying points in limited numbers; however, when there are too many outliers, especially in large highly structured or multivariate datasets, detection and handling of outliers becomes difficult or even unfeasible~\cite{El-Melegy2009}. Using a loss function robust to outliers could mitigate the need to clean training data.

\begin{figure}[htb!]
	\centering
	\includegraphics[width=0.6\textwidth]{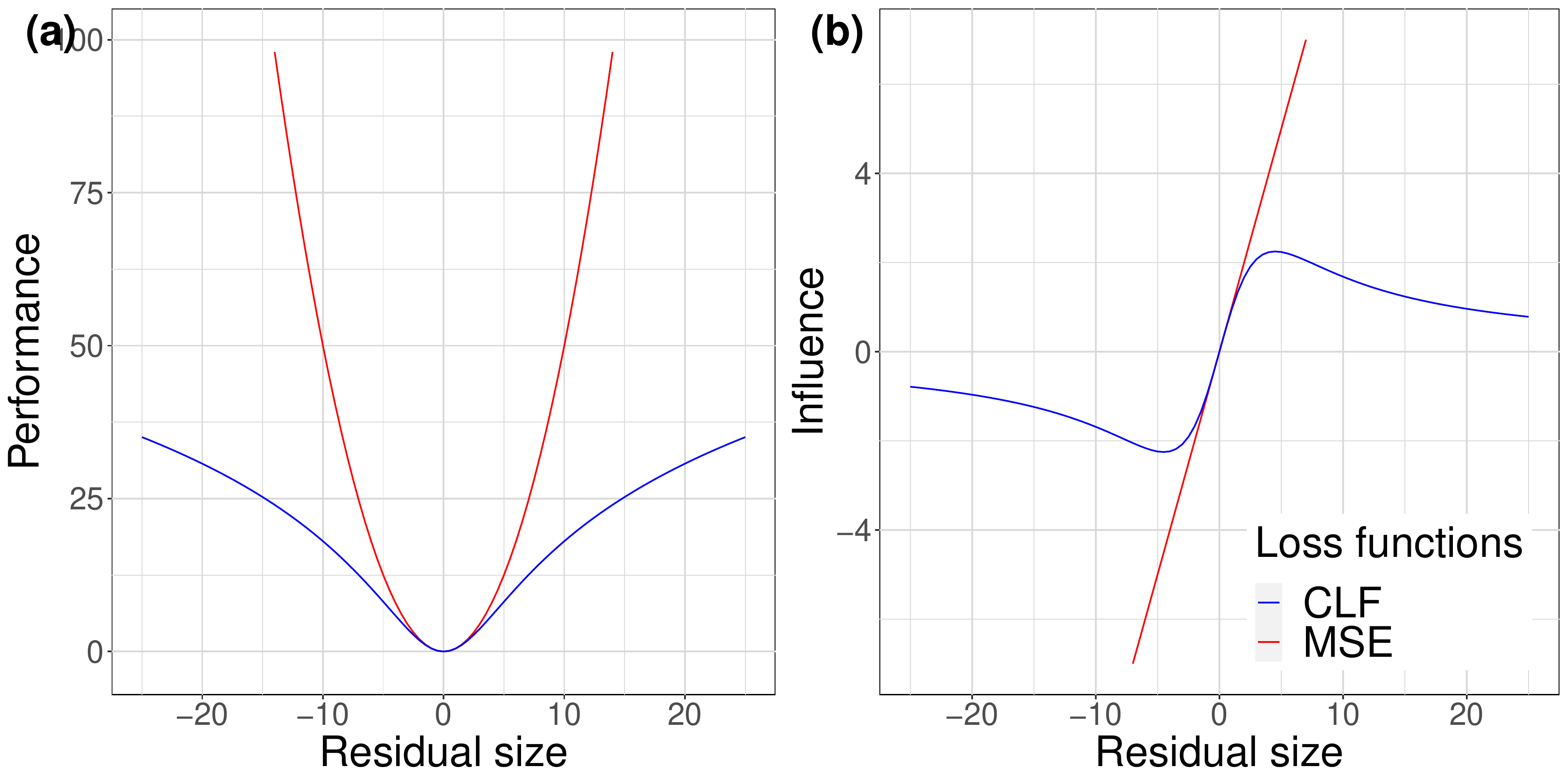}
	\caption{\textbf{(a)} Quantification of performance according to MSE and CLF varying with with residual size, \textbf{(b)} Influence on performance quantification varying with residual size.}
    \label{fig:MSE_vs_CLF_infl}
\end{figure}

Hence, alternative loss functions for regression have been used in attempts to achieve robustness to outliers and non-Gaussian noise profiles. Particularly, M-estimators based on the maximum likelihood for the density functions of an assumed distribution are common candidates \cite{Brunet2010}. The Cauchy loss function (CLF) is one popular choice that assumes a Cauchy noise profile \cite{Zahra2014}. The Cauchy distribution
has statistical properties that align closely with the intuition of what an outlier might be, such as its undefined mean and infinite variance, accounting for more extreme residuals than would be expected under a Gaussian assumption \cite{Borak2005}.
The most convincing 
argument for the CLF is that the influence that residuals have on the estimator is bounded and tends towards zero (illustrated in Fig. \ref{fig:MSE_vs_CLF_infl}) \cite{Li2019}.

Despite some theoretical and empirical evidence of the advantages of using CLF over MSE \cite{El-Melegy2009}, there is a lack of corresponding systematic studies that would highlight the behaviour of the two loss functions under different parameter settings. 
Thus, the research objectives of this study are summarised as follows: 
\begin{itemize}
    \item To establish whether the use of CLF consistently leads to similar or better performance of the artificial neural network (ANN) models compared to MSE, applied to regression problems with varying degrees of Gaussian and Cauchy noise.
    \item To elucidate whether the use of CLF in place of MSE yields ANN models that are more robust in the presence of deterministic noise, arising from complexities in a regression dataset that a model has limited flexibility to capture otherwise.
    \item To contrast the behaviour and robustness of ANN models trained with CLF to those trained with MSE in the presence of outliers in real world data.
\end{itemize}

The rest of the paper is structured as follows: 
Section~\ref{sec:bg} covers the relevant background to contextualise the research.  Section~\ref{sec:method} outlines the methodology that is used in addressing the research goals. Section~\ref{sec:discuss} then presents and discusses the results of the research, and conclusions are drawn in Section~\ref{sec:conclude}.

\section{Background}\label{sec:bg}
This section provides background into MSE, CLF, and their statistical considerations. Furthermore, a brief discussion is offered on deterministic noise, and an argument is made for the validity of the inferences drawn from the results of the experiments.

\subsection{Consequences of the Gaussian assumption}
The normal or Gaussian distribution is arguably the most extensively used distribution in most fields of science \cite{Park2013}. In particular, the noise found in empirical data for many scientific domains is modelled as or assumed to be Gaussian \cite{Park2013}, based on the statistical central limit theorem and the law of large numbers, which assume a number of regularity conditions \cite{Borak2005}. Furthermore, a number of advantageous properties regarding unbiased parameter estimates and least squares methods come under the assumption of Gaussianity~\cite{Pearson2003}. Namely, the maximum likelihood estimator is equivalent to the least squares estimator for data containing Gaussian noise \cite{Brunet2010}.

As it applies to ANNs, the minimisation of the MSE loss can be shown to be equivalent to the selection of parameters that maximise the likelihood for a Gaussian model, i.e. its maximum likelihood estimator \cite{Chambers2012}. 
Hence, the convention of Gaussian noise is subsumed in optimisation of the MSE loss for regression problems \cite{Heravi2018,Chen2018}.

However, this conventional assumption of Gaussianity is often unfounded in practice, leading to significant degradation in results in the presence of adversarial, impulsive noise components, and particularly in the case of non-Gaussian noise \cite{Tsakalides1994}. A fundamental error is often made in the use of the central limit theorem, where finite variances are assumed although this may not be appropriate \cite{Brunet2010}. In these cases, the MSE loss breaks down, and the estimator vector may shift significantly from the true regression plane to minimise the errors that are inflated by being squared \cite{Chen2018}. Moreover, the influence of these outliers on the overall estimate increases proportionally with their magnitude in an unbounded manner \cite{Li2019,Barron2019}. These characteristics have led practitioners to seek performance criteria that are more robust to outliers and non-Gaussian noise profiles.

\subsection{Stable distributions}
Although there exist a number of distributions accounting for large residuals (i.e. having heavier tails) than the Gaussian distribution, Borak et al. \cite{Borak2005} express a theoretical reasoning for assessing the use of the family of stable distributions in particular. A generalisation of the central limit theorem states that stable distributions ``are the only possible limit distributions for properly normalised and centred sums of independent, identically distributed random variables'' \cite{Borak2005}. This statement of the central limit theorem drops the regularity conditions (such as finite variance) necessary to converge to a Gaussian, and even subsumes the convergence to a Gaussian as a member of the family of stable distributions. Hence, the use of a heavy-tailed stable distribution offers a correction to the frequent misapplication of the central limit theorem that accompanies the conventional assumption of Gaussianity.
Although there exist an infinite number of stable distributions, there exist only two closed form formulas for symmetric stable distributions: the Gaussian and Cauchy distributions \cite{Borak2005}.

\subsection{Cauchy distribution and Cauchy loss function}

The choice of the Cauchy distribution is more intuitive for modelling practical occurrences of noise and outliers than the Gaussian. Its close-form density function also makes Cauchy distribution a candidate for maximum likelihood estimation \cite{Borak2005}, and hence its use 
as a loss function based on a heavy-tailed stable distribution. This paper considered CLF scaled by a constant as presented in \cite{Zahra2014} and shown as necessary to achieve the theoretical behaviour of CLF in preliminary testing:

$$L_{CLF} = \frac{c^2}{2} \log\left(1 + \left(\frac{y - \hat{y}}{c} \right)^2 \right)$$
where $c\in(0, \infty)$ is the CLF constant, and $y$ and $\hat{y}$ refer to the target and actual output of the model, respectively, their difference constituting the residual.
The influence function associated with the CLF indicates any residual’s influence on the estimate has an upper bound, and that for arbitrarily big residuals, their influence tends towards zero \cite{Li2019}. This is as a result of the Cauchy distribution’s power-law behaviour \cite{Borak2005}, and qualifies as robust to noise when contrasted to MSE \cite{Li2019}.

\subsection{Deterministic noise}
Although the influence of additive noise, whether Gaussian or Cauchy, has been the primary grounds for discussion until this point, a discussion on the tolerance of the MSE and CLF performance criteria would be incomplete without consideration for the internal complexities of a dataset. In the field of Bayesian statistical models, a given signal from a random process can be decomposed into a structured deterministic component (explained by the model) and a random stochastic component (attributed to noise) \cite{Huang2000}. 

An analogue is drawn between the Bayesian model and a supervised learning algorithm by Abu-Mostafa et al. \cite{Abu-Mostafa2012}, who assert that a well-performing model not only accounts for random stochastic noise (in the case of this paper assumed to be either Gaussian or Cauchy), but also for complexity in the structured deterministic component. Complexity that is not modelled is treated as deterministic noise, which a model may fail to fit, even in the absence of stochastic noise \cite{Abu-Mostafa2012}. Ultimately, the performance of models using MSE or CLF are subject not only to the noise present in the data, but also whether or not they offer the degrees of freedom to capture the deterministic complexities of the target function.

\subsection{On the validity of inferring from the results}
The performance of statistical inference based on the samples generated from the experiments of this study warrants careful consideration, particularly the results generated in the presence of Cauchy noise. 
The assumption of Cauchy noise implies indeterminate moments \cite{Borak2005}. In particular, Cauchy population mean is undefined, and its variance is infinite \cite{Borak2005}. Consequently, inference that depends on any assumption of convergence to a finite population variance or determinate population mean can be thought to be limited in usefulness. Thus, inferences based on the root mean squared error (RMSE) are likely to be unreliable in the presence of Cauchy noise.

An empirical study conducted by Balkema and Embrechts~\cite{Balkema2018} aimed to assess the performance of various estimators in a simple linear regression problem with deterministic and stochastic components, both modelled as potentially heavy tailed random variables. Balkema and Embrechts showed that under the condition that the deterministic component 
is of finite variance itself, the least absolute deviation (i.e. the minimisation of the mean absolute error (MAE)) is typically a good estimator for the linear regression parameters, even if the stochastic component has infinite variance \cite{Balkema2018}. In the context of stochastic gradient descent, the minimisation of MAE is equivalent to the least absolute deviation. Hence, by assuming finite variance of the multivariate random process giving rise to the data set, 
the use of MAE is considered to be fairly reliable \cite{Balkema2018}.


The inferential tools that were chosen in this study (the Kruskal-Wallis \cite{Fan2010} and Wilcoxon Rank Sum \cite{Brcich2005} tests) are rank-based tests to avoid mean-based testing. They are hence equivalent to median-based testing when the various inferential population distributions (i.e., the actual performance of the various models) are assumed to be of similar shapes \cite{Hart2001}. 

\subsection{Related Work}\label{sec:related}

The CLF has been applied and achieved satisfactory results in ANNs as well as other machine learning domains such as subspace clustering \cite{El-Melegy2009,Li2019,Barron2019}. It is suggested by Li et al. \cite{Li2019} that the nature of the CLF’s influence function means that it is generally insensitive to the distribution of the noise; however, developing a better theoretical understanding is still worthwhile. Furthermore, El-Melegy et al. \cite{El-Melegy2009} suggest that CLF may only be robust to a certain extent. As such, a comparison of how CLF and MSE perform in the presence of varying degrees of both Gaussian and non-Gaussian stable noise is necessary in order to evaluate CLF’s ability to generalise, and its robustness to the implied Cauchy noise profile. 

Additionally, a gap in the literature was found regarding the impact of deterministic noise on the performance of CLF models, and how this compares to the performance of MSE models. Lastly, none of the literature surveyed exhibited a statistically sensitive approach to the drawing of conclusions.

\section{Methodology}\label{sec:method}
Three experiments were conducted for this study. The first two used handcrafted datasets of varied complexity to assess the responses of MSE and CLF to various Gaussian and Cauchy noise profiles. The third experiment made use of a real-world dataset to assess the response of the two loss functions to simulated outliers. This section describes each of the datasets, the structural and implementational considerations of each experiment, and the corresponding experimentation procedures.

\subsection{Handcrafted experiments}
Two regression datasets were generated according to mathematical functions. The first dataset is in two variables, given by:
\begin{equation}y_1 = e^{x_1} - \sin{x_2}\end{equation}
for $x_1\in [-6, 2]$ and $x_2 \in [-3,9]$.

The second dataset is generated from  a function of eight variables, characterised by quickly oscillating behaviour. It is given by:

\begin{equation}
\begin{split}
y_2 = &0.03\Big( \sin^2{(x_1)}(x_2-2)(x_3-8)(x_4-11) +\\ 
&\cos^2{(x_5)}(x_6-6)(x_7-6)(x_8+5)^2 \Big)
\end{split}
\end{equation}
for $x_1\in [-6,17]$, $x_2\in [-7, 20]$, $x_3\in [-2, 17]$, $x_4\in [-6, 10]$, $x_5\in [-10,16]$, $x_6\in [-5,10]$, $x_7\in [-15, 9]$, and  $x_8\in [-1, 14]$.

Each dataset consisted of 5000 randomly sampled datapoints, where the target values followed the given mathematical function $y_1$ or $y_2$. 

In the experimentation, noise was introduced into each dataset for training purposes, and the integrity of the data was maintained for testing. For the Gaussian configuration of each experiment, additive noise $\epsilon$ was injected into $y_i$, where $\epsilon \sim N(0, \sigma^2)$, given that the standard deviation $\sigma \in \{1, 10, 50, 100\}$.

Likewise, for the Cauchy configuration, additive noise $\epsilon$ was injected into $y_i$ where $\epsilon \sim Cau(x_0, \tau)$, given that the location parameter $x_0 = 0$ and the scale parameter $\tau \in \{1, 10, 50, 100\}$.

\subsection{Seoul bike sharing demand experiment}

The dataset is a compilation of a year’s worth of data from the Seoul Public Data Park website. Each datapoint describes the number of bikes rented in the city for a single hour of a particular day, along with a number of meteorological and date-specific details that may have correlated with the demand for bikes  \cite{Sathishkumar2020}. This dataset is available on the UC Irvine Machine Learning Repository. 

In the experimentation, noise was injected according to a uniform distribution with a range that is 500 times the range of the actual data. This was done in order to simulate outliers. Noise was randomly injected into variable proportions of the training data targets according to the experiment configuration (i.e. 2.5\%, 5\%, 7.5\%, 10\% of the training data). The testing set was left uncorrupted.

It is noted by Qi et al. \cite{Qi2021} that data from the UC Irvine Machine Learning Repository is typically clean, so inconsistencies are introduced here in the form of simulated outliers. This noise is essentially impulsive, as the uniform distribution acts as a thick-tailed distribution along its range, and was picked as an outlier generator on this basis.

\subsection{General setup}
For each experiment, there were two model types used. Each type is topographically analogous: a feed-forward ANN with the Adam optimiser \cite{Zhang2018} and the ReLU activation function \cite{Banerjee2019} applied in the hidden layers. For the handcrafted experiments, the ANN architecture featured a single hidden layer of ten nodes, and a single output node. For the Seoul bike sharing demand experiment, a bigger ANN architecture was used, featuring two hidden layers of fourteen nodes each, and a single output node. The ANN architecture parameters were not optimised for a specific loss function to avoid bias.

For each experiment, the models differed only in the loss function employed. The first model type makes use of MSE, and the second makes use of CLF\textsubscript{c}, where $c$ is the CLF constant. For the handcrafted experiments, five CLF\textsubscript{c} models were considered for $c \in \{0.1, 1, 10, 20, 100 \}$. For the Seoul bike sharing demand experiment, six CLF\textsubscript{c} models were considered for $c \in \{1, 10, 100, 200, 1 000, 10 000 \}$.

All experiments were conducted on Google Colab, using the Python 3 Google Compute Engine backend. The setup features 12.78GB memory and two Intel® Xeon® 2.30GHz CPUs. All code was written in Python, using algorithms provided by the open-source \textit{Keras}, \textit{TensorFlow} and \textit{scikit-learn} libraries. As no CLF implementation exists in any of the used libraries, a custom implementation was written for the purposes of the research.

\subsection{General procedure}
In all experiments, 10-fold cross-validation was used to split the datasets into training and testing sets. As a negative control and benchmark, training and testing were initially performed on uncorrupted data using both the MSE model and all CLF\textsubscript{c} models for each of the datasets. 

Subsequently, training was performed on the dataset in question with corrupted data. In the handcrafted experiments, this means corrupting the data with either Gaussian or Cauchy noise. Four extents of Gaussian noise were assessed, where the noise profile had a standard deviation $\sigma \in \{1, 10, 50, 100\}$. Likewise, four extents of Cauchy noise were assessed, where the noise profile had a scale parameter $\tau \in \{1, 10, 50, 100\}$. The training was performed for the MSE model and all CLF\textsubscript{c} models, which were then tested on uncorrupted data.

In the Seoul bike sharing demand experiment, training data was corrupted with simulated outliers, generated according to a uniform distribution. The proportion of the dataset to be corrupted was set to one of four levels: 2.5\%, 5\%, 7.5\% or 10\%. The training was performed for the MSE model and all CLF\textsubscript{c} models, which were then tested on uncorrupted data.

To score the models' performance on the test data, two different metrics were used for all experiments: RMSE and MAE.
Five values of each score were generated from five independent training-testing runs for each model,  i.e., each value is the average of the 10 scores obtained from each cross-validation process.

\section{Discussion}\label{sec:discuss}

The results from the experimental procedures are presented and discussed in this section. The 2-variable and 8-variable handcrafted experiments are presented first, followed by the results from the Seoul bike sharing demand experiment. 

\subsection{2-variable handcrafted experiment}
Tables~\ref{table:HCTable_Control_Gaussian} and~\ref{table:HCTable_Cauchy} summarise the MSE and CLF$_c$ performance under various extents of Gaussian and Cauchy noise, respectively.

\begin{table*}[h] 
\centering
\scriptsize
\setlength{\tabcolsep}{0.3pt}
\caption{Handcrafted experiments: negative control and Gaussian noise. Lowest errors for each set-up are indicated in red.}
\label{table:HCTable_Control_Gaussian}
\begin{tabular}{|c  c | c c | c c | c c | }
\hline
  &  & \multicolumn{2}{|c|}{Negative} & \multicolumn{2}{|c|}{$\sigma =1$} & \multicolumn{2}{|c|}{$\sigma =10$} \\
\hline
Var & Model & MAE & RMSE & MAE & RMSE &  MAE & RMSE \\
\hline
2 & CLF\textsubscript{0.1} & 0.485 (0.013) & 0.67 (0.006) & 0.492 (0.021) & 0.675 (0.031) & 0.617 (0.021) & 0.8 (0.026) \\
&CLF\textsubscript{1.0} & \textbf{\textcolor{red}{0.476 (0.012)}} & 0.663 (0.008) & 0.491 (0.006) & 0.666 (0.012) & 0.614 (0.013) & 0.799 (0.019) \\
&CLF\textsubscript{10.0} & 0.479 (0.009) & \textbf{\textcolor{red}{0.657 (0.008)}} & \textbf{\textcolor{red}{0.487 (0.009)}} & \textbf{\textcolor{red}{0.664 (0.014)}} & 0.61 (0.016) & 0.79 (0.015) \\
&CLF\textsubscript{20.0} & 0.482 (0.01) & 0.668 (0.007) & 0.495 (0.008) & 0.672 (0.006) & \textbf{\textcolor{red}{0.604 (0.007)}} & 0.795 (0.005) \\
&CLF\textsubscript{100.0} & 0.48 (0.012) & 0.659 (0.012)  & 0.494 (0.008) & 0.674 (0.008)  & 0.653 (0.019) & 0.841 (0.01)  \\
& MSE & 0.485 (0.01) & 0.677 (0.013)  & 0.496 (0.009) & 0.678 (0.008)  & \textbf{\textcolor{red}{0.604 (0.01)}} & \textbf{\textcolor{red}{0.785 (0.01)}}  \\
\hline
8 &CLF\textsubscript{0.1} & \textbf{\textcolor{red}{39.261 (0.059)}} & 55.384 (0.155) & 39.383 (0.188) & 55.607 (0.359)  & 39.075 (0.033) & 54.881 (0.06) \\
&CLF\textsubscript{1.0} & 39.573 (0.485) & 55.91 (0.839) & \textbf{\textcolor{red}{39.195 (0.053)}} & 55.255 (0.134) & 39.059 (0.041) & 54.859 (0.074) \\
&CLF\textsubscript{10.0} & 39.351 (0.234) & 55.533 (0.439) & 39.424 (0.318) & 55.703 (0.568)  & 39.073 (0.035) & 54.885 (0.085) \\
&CLF\textsubscript{20.0} & 39.445 (0.152) & 55.709 (0.244) & 39.306 (0.134) & 55.478 (0.303)& \textbf{\textcolor{red}{39.043 (0.05)}} & 54.851 (0.084)  \\
&CLF\textsubscript{100.0} & 39.394 (0.17) & 55.649 (0.314) & 39.572 (0.457) & 55.917 (0.815)& 39.063 (0.04) & 54.848 (0.068) \\
&MSE & 41.81 (0.073) & \textbf{\textcolor{red}{52.425 (0.028)}} & 41.749 (0.074) & \textbf{\textcolor{red}{52.398 (0.042)}} & 41.822 (0.085) & \textbf{\textcolor{red}{52.426 (0.024)}} \\
\hline
  &  & \multicolumn{2}{|c|}{$\sigma =50$} & \multicolumn{2}{|c|}{$\sigma =100$}\\
\cline{1-6}
2 & CLF\textsubscript{0.1} & 1.189 (0.03) & 1.531 (0.043)  & 1.873 (0.502) & 2.34 (0.51) \\
&CLF\textsubscript{1.0} & 1.106 (0.132) & 1.454 (0.151)  & 1.655 (0.181) & 2.056 (0.164) \\
&CLF\textsubscript{10.0} &  1.217 (0.082) & 1.555 (0.125)& 1.865 (0.373) & 2.341 (0.368) \\
&CLF\textsubscript{20.0} &  1.216 (0.124) & 1.535 (0.108)  & 1.803 (0.385) & 2.266 (0.431) \\
&CLF\textsubscript{100.0} & 1.401 (0.176) & 1.784 (0.228)  & 1.732 (0.441) & 2.173 (0.514) \\
& MSE &  \textbf{\textcolor{red}{0.994 (0.156)}} & \textbf{\textcolor{red}{1.296 (0.18)}}  & \textbf{\textcolor{red}{1.376 (0.174)}} & \textbf{\textcolor{red}{1.729 (0.171)}} \\
\cline{1-6}
8 & CLF\textsubscript{0.1} & 40.113 (0.179) & 53.442 (0.142)  & 41.686 (0.651) & 55.161 (0.714) \\
&CLF\textsubscript{1.0} &  40.305 (0.26) & 53.454 (0.192) & 41.975 (0.526) & 54.97 (0.856) \\
&CLF\textsubscript{10.0} &  40.033 (0.114) & 53.359 (0.084) & 42.051 (0.508) & 54.837 (0.597) \\
&CLF\textsubscript{20.0} &  40.45 (0.39) & 53.721 (0.352)  & 42.185 (0.73) & 54.642 (0.581) \\
&CLF\textsubscript{100.0} & \textbf{\textcolor{red}{40.023 (0.032)}} & 53.497 (0.221) & \textbf{\textcolor{red}{41.669 (0.411)}} & 54.18 (0.44)\\
& MSE & 41.803 (0.087) & \textbf{\textcolor{red}{52.438 (0.038)}}  & 41.84 (0.196) & \textbf{\textcolor{red}{52.504 (0.036)}} \\
\cline{1-6}
\end{tabular}
\end{table*}

\begin{table*}[h] 
\centering
\scriptsize
\caption{Handcrafted experiments: Cauchy noise.  Lowest errors for each set-up are indicated in red.}
\label{table:HCTable_Cauchy}
\begin{tabular}{|c  c | c c | c c | c c | c c |}
\hline
  &  & \multicolumn{2}{|c|}{$\tau =1$} & \multicolumn{2}{|c|}{$\tau =10$} \\
\hline
Var & Model & MAE & RMSE &  MAE & RMSE \\
 \hline
 2 & CLF\textsubscript{0.1} & 0.503 (0.011) & \textbf{\textcolor{red}{0.68 (0.014)}} & 0.655 (0.025) & 0.867 (0.029)  \\
 & CLF\textsubscript{1.0} & 0.506 (0.009) & 0.685 (0.005)  & 0.64 (0.018) & 0.847 (0.025)   \\
 & CLF\textsubscript{10.0} & 0.517 (0.009) & 0.693 (0.006) & \textbf{\textcolor{red}{0.635 (0.01)}} & \textbf{\textcolor{red}{0.836 (0.015)}} \\
 &CLF\textsubscript{20.0} & 0.503 (0.006) & 0.69 (0.011) & 0.645 (0.03) & 0.85 (0.027)  \\
 &CLF\textsubscript{100.0} & \textbf{\textcolor{red}{0.5 (0.011)} }& 0.689 (0.016)  & 0.659 (0.017) & 0.861 (0.036)  \\
 & MSE & 1.088 (0.11) & 1.377 (0.111)   & 1.952 (0.22) & 2.426 (0.223)  \\
\hline
8 & CLF\textsubscript{0.1} & 39.289 (0.125) & 55.446 (0.323)  & 39.028 (0.016) & 54.786 (0.093)   \\
&CLF\textsubscript{1.0} & 39.347 (0.113) & 55.511 (0.23) & 38.99 (0.038) & \textbf{\textcolor{red}{54.669 (0.084)}} \\
&CLF\textsubscript{10.0} & 39.507 (0.459) & 55.848 (0.916)  & 39.118 (0.164) & 54.939 (0.318) \\
&CLF\textsubscript{20.0} & \textbf{\textcolor{red}{39.194 (0.089)}} & 55.236 (0.191) & 39.179 (0.139) & 55.066 (0.237)  \\
&CLF\textsubscript{100.0} & 39.206 (0.061) & 55.251 (0.13)   & \textbf{\textcolor{red}{39.017 (0.033)}} & 54.726 (0.071) \\
&MSE & 41.772 (0.099) & \textbf{\textcolor{red}{52.531 (0.127)}}   & 45.717 (2.667) & 60.302 (4.179)   \\
\hline
  &  & \multicolumn{2}{|c|}{$\tau =50$} & \multicolumn{2}{|c|}{$\tau =100$}\\
\hline
 2 & CLF\textsubscript{0.1}& \textbf{\textcolor{red}{1.289 (0.147)}} & \textbf{\textcolor{red}{1.654 (0.146)}} & 1.937 (0.254) & 2.392 (0.266)\\
 & CLF\textsubscript{1.0}& 1.512 (0.131) & 1.931 (0.167) & \textbf{\textcolor{red}{1.595 (0.228)}} & \textbf{\textcolor{red}{2.007 (0.222)}}\\
 & CLF\textsubscript{10.0} & 1.319 (0.209) & 1.669 (0.239)   & 1.769 (0.334) & 2.282 (0.384)\\ 
 &CLF\textsubscript{20.0} & 1.365 (0.181) & 1.772 (0.168)  & 1.669 (0.32) & 2.082 (0.352) \\
 &CLF\textsubscript{100.0} & 1.337 (0.118) & 1.749 (0.12)  & 1.847 (0.318) & 2.255 (0.272) \\
 & MSE& 2.031 (0.436) & 2.494 (0.533)  & 1.85 (0.318) & 2.305 (0.318)  \\
 \hline
 8 & CLF\textsubscript{0.1}& 40.085 (0.176) & 54.368 (0.338)  & 42.112 (0.44) & 56.261 (0.502) \\
 & CLF\textsubscript{1.0}& 39.941 (0.195) & 54.237 (0.385)  & 42.515 (0.697) & 56.806 (0.859) \\
 & CLF\textsubscript{10.0} & 39.95 (0.327) & \textbf{\textcolor{red}{53.966 (0.313)}}  & 42.373 (0.474) & 56.966 (0.788) \\
 &CLF\textsubscript{20.0} & \textbf{\textcolor{red}{39.866 (0.244)}} & 54.171 (0.268) & 42.092 (0.547) & 56.938 (0.549) \\
 &CLF\textsubscript{100.0}  & 39.888 (0.118) & 54.194 (0.247) & \textbf{\textcolor{red}{41.853 (0.653)}} & \textbf{\textcolor{red}{55.916 (0.516)}} \\
 & MSE& 50.987 (1.734) & 71.354 (2.23)  & 53.683 (0.664) & 75.103 (0.987)\\ \hline
\end{tabular}
\end{table*}

The negative control configuration (refer to Table~\ref{table:HCTable_Control_Gaussian}) saw no additive noise in the data, and the performances of the six models were comparable, with no conclusive evidence suggesting significant differences between them. This behaviour was as expected, indicating that CLF is at the least not inferior to MSE.\looseness=-1 

\begin{figure}[tb!]
    \centering
    \caption{MAE scores for models varying with standard deviation $\sigma$ of Gaussian noise  and scale parameter $\tau$ of Cauchy noise (2-variable handcrafted experiment)}
        \subfloat[Gaussian]{\includegraphics[width=0.4\columnwidth]{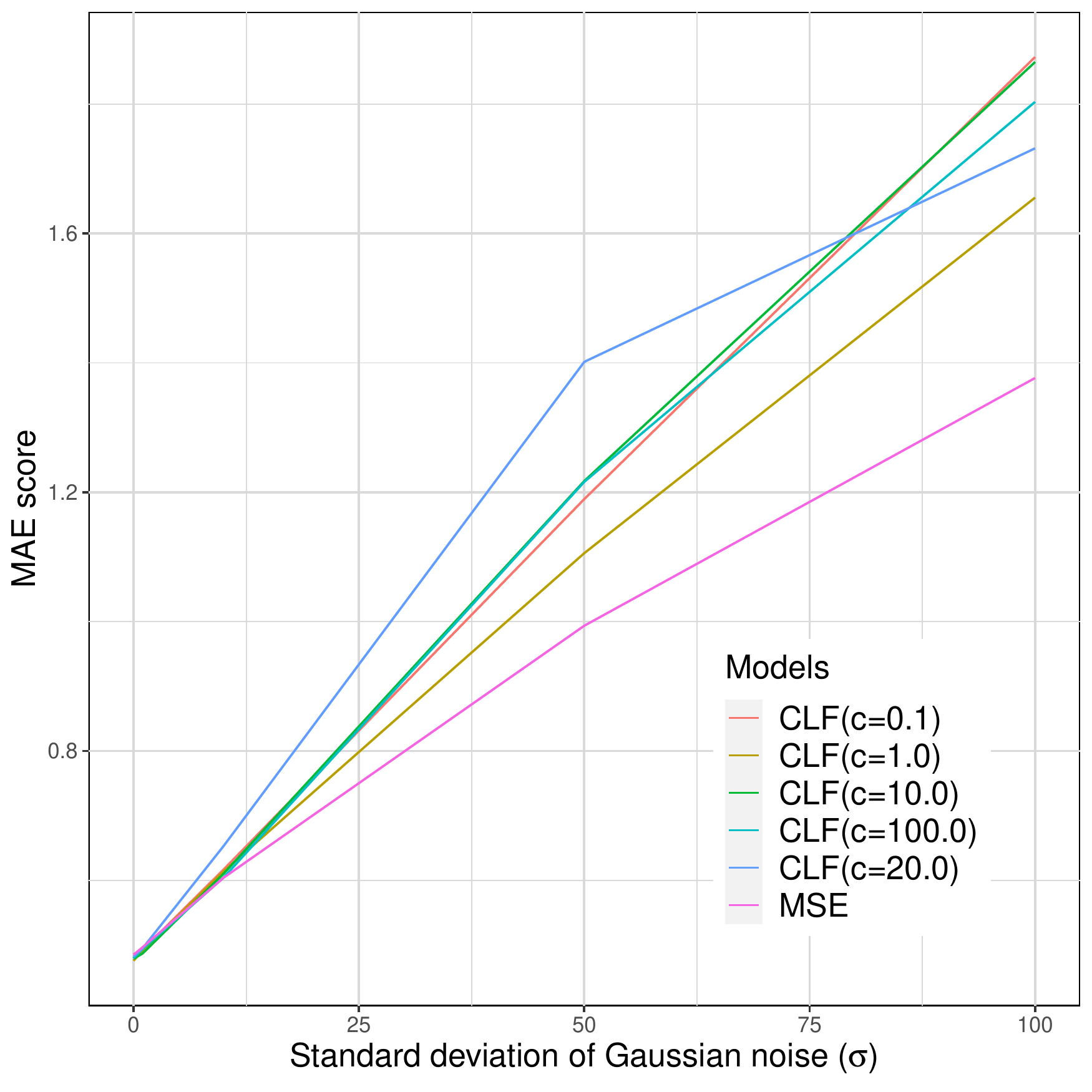}}\label{fig:HC2_Gaus_MAE}
        \qquad
        \subfloat[Cauchy]{\includegraphics[width=0.4\columnwidth]{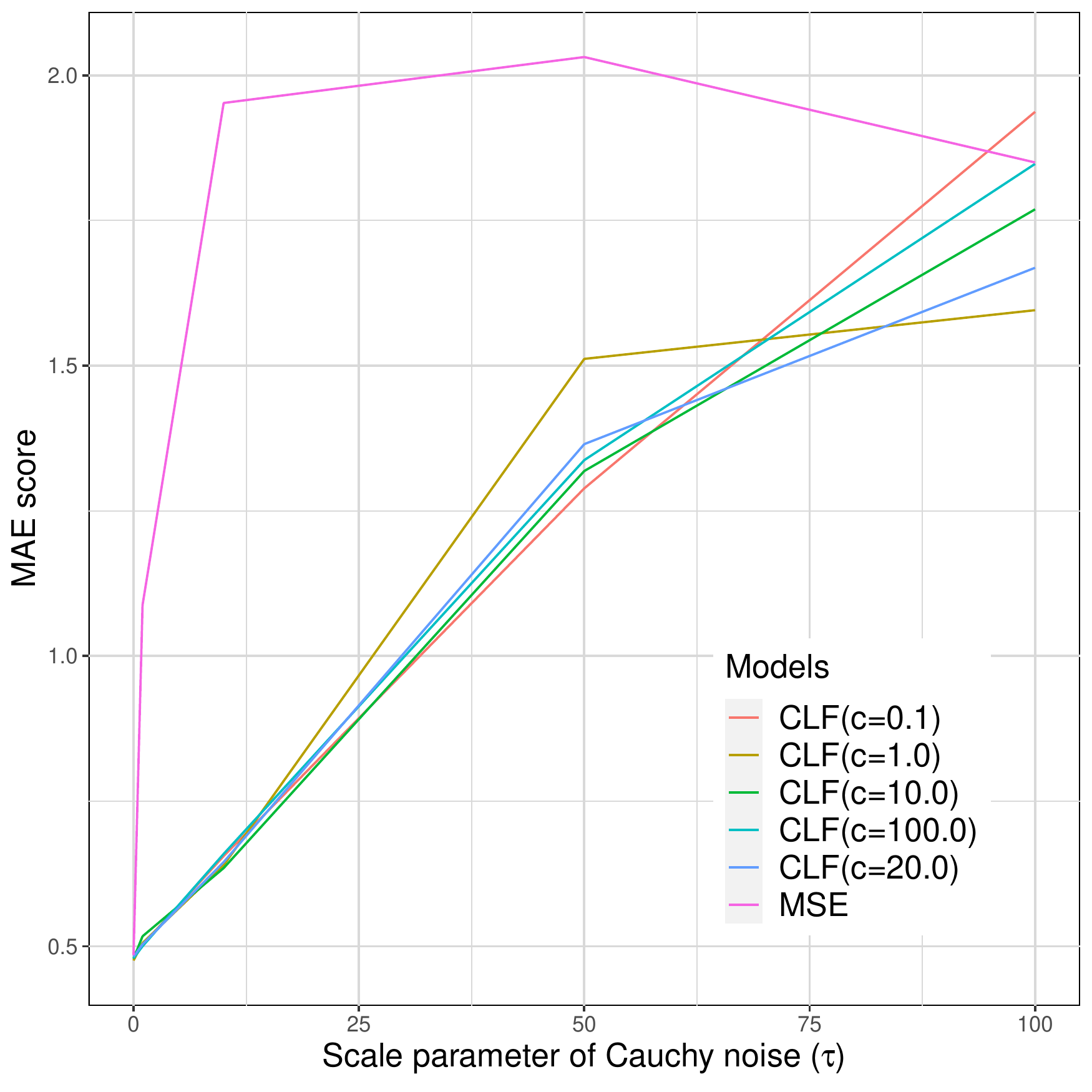}}\label{fig:HC2_Cauc_MAE}
    \label{fig:subfigname_name}
\end{figure}

The experiment configuration that used Gaussian noise profiles also exhibited expected behaviour. Increasing the standard deviation of the Gaussian noise resulted in a fairly steady decline in performance in all models (see Fig.~\ref{fig:HC2_Gaus_MAE}). However, the CLF models did not consistently perform equivalently to the MSE model. From the point where $\sigma=10$, WRS tests revealed evidence  ($p<0.01$ for both scores) of notable difference between the other models and the CLF\textsubscript{20} model. 
By the point where $\sigma=50$, all CLF models have performed worse than the MSE model according to KW and WRS testing ($p<0.05$ for both scores). 
The superior performance of the MSE model on its loss function's assumed noise profile is to be expected, and ultimately the rates of degradation of CLF models' performance are comparable for limited extents of Gaussian noise. 

However, 
beyond $\sigma =10$, the CLF\textsubscript{1} model was shown to perform better than the worst performing CLF models in each case, according to evidence from WRS testing ($p<0.05$ in both scores). This is illustrated in Fig.~\ref{fig:HC2_Gaus_MAE}, and may speak to the optimisation obtained from choosing the appropriate CLF constant, noting that worse performance was seen for both smaller and larger CLF constants.


For the Cauchy noise configuration, the results were as expected once more (refer to Table~\ref{table:HCTable_Cauchy}). Noting that inference can only be thought to be statistically sound for the MAE scores, these form the basis of discussion. From $\tau=1$ and until $\tau=100$, the MSE model was outperformed by all CLF models according to evidence from WRS and KW testing ($p<0.005$ for MAE scores). The MSE model had test MAE scores that reflected consistently poor performance, as seen in Fig.~\ref{fig:HC2_Cauc_MAE}.\looseness=-1

In contrast, the CLF models' performances degraded gradually with increasing $\tau$. 
However, by $\tau=100$, the performance degradation of most CLF models led to insufficient evidence of their being different from the MSE model, according to KW testing.

Furthermore, WRS testing confirmed ($p<0.05$ in both scores) that CLF models performed differently from each other.
Thus, the results showed 
that in the presence of all the considered extents of additive noise, at least one CLF model outperformed the MSE model. This indicates that there may be some correlation between the scale of Cauchy noise profiles and optimal value of the CLF constant.

\subsection{8-variable handcrafted experiment}

The negative control configuration saw no additive noise in the data (see Table~\ref{table:HCTable_Control_Gaussian}). The results obtained from this benchmark present a seeming contradiction in the scores. According to the MAE score, KW and WRS testing reflected significant evidence ($p<0.005$) of the CLF models outperforming the MSE model. However, according to the RMSE score, KW and WRS testing reported equally significant evidence ($p<0.005$) of the MSE model performing better than any of the CLF models. It is likely that this apparent contradiction is a consequence of deterministic noise. 
Nonetheless, WRS testing asserted further that there was evidence ($p<0.05$ in both scores) of different performances between the CLF models, indicating that the performance of CLF is dependent on the CLF constant choice. 
An example of this is in the CLF\textsubscript{0.1} model performing better than the CLF\textsubscript{20} model. 



\begin{figure}[htb!]
    \centering
    \caption{MAE and RMSE scores for models varying with standard deviation $\sigma$ of Gaussian noise (8-variable handcrafted experiment)}
        \subfloat[MAE]{\includegraphics[width=0.4\columnwidth]{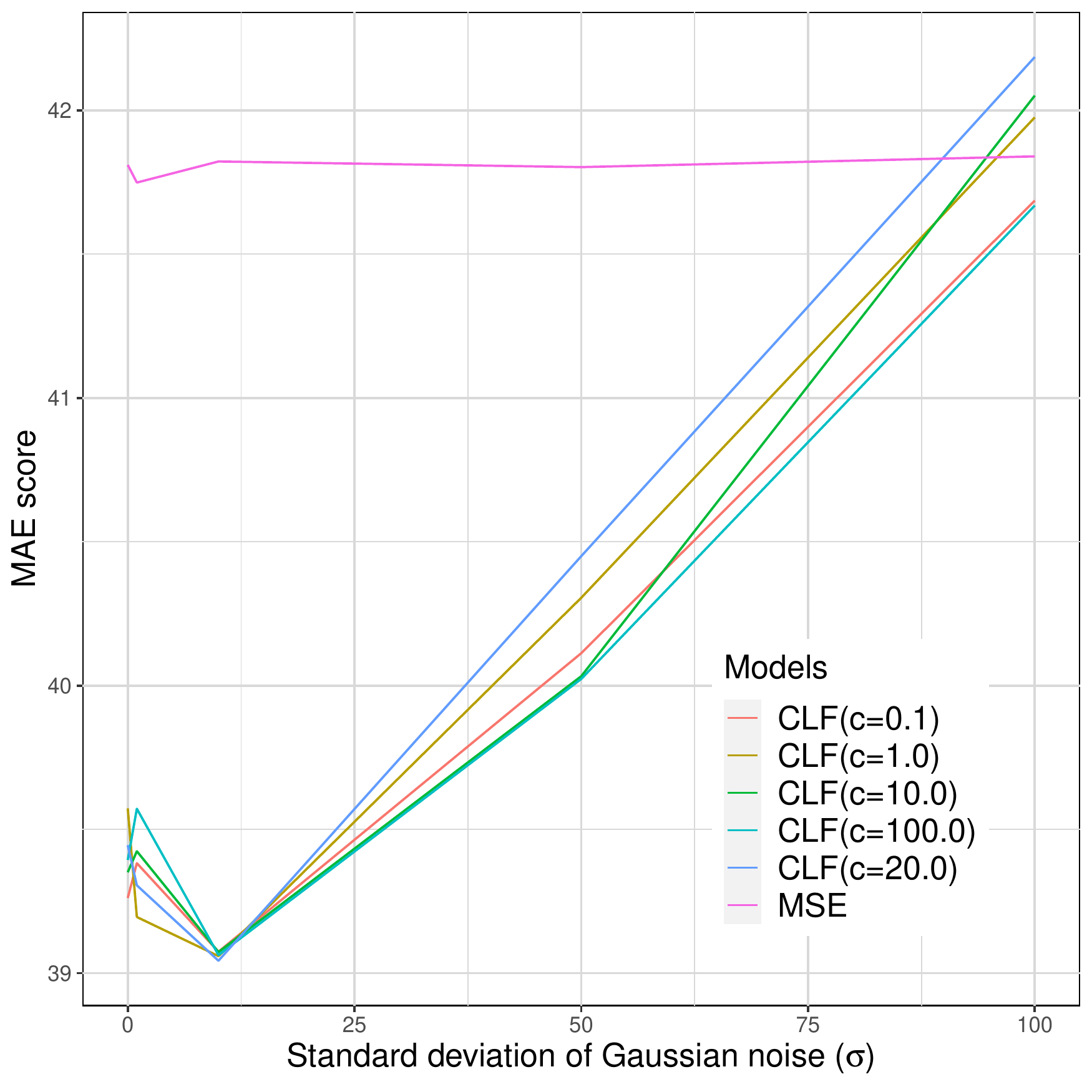}}\label{fig:HC8_Gaus_MAE}
        \qquad
        \subfloat[RMSE]{\includegraphics[width=0.4\columnwidth]{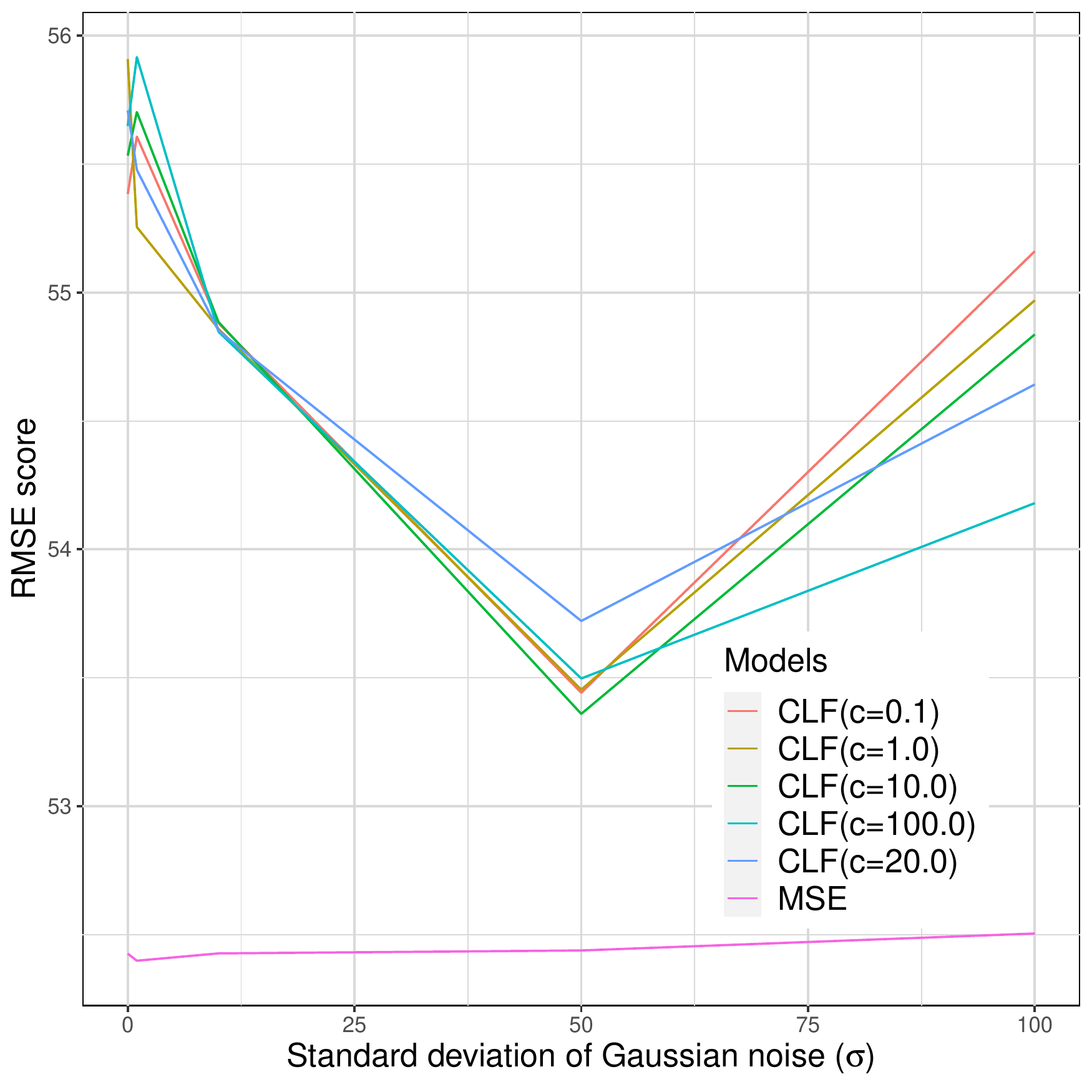}}\label{fig:HC8_Gaus_RMSE}
    \label{fig:subfigname2}
\end{figure}

As was the case for the negative control, there was an apparent contradiction in the scores for the various models under Gaussian noise (see Table~\ref{table:HCTable_Control_Gaussian}). The MSE model could be shown to perform better than the CLF models in the RMSE scores for all considered values of $\sigma$ (refer to Fig.~\ref{fig:HC8_Gaus_RMSE}). Similarly, the CLF models could be shown to perform better in the MAE scores  (as seen in Fig.~\ref{fig:HC8_Gaus_MAE}). Each contradicting case of better performance was supported by statistical evidence ($p<0.05$) through KW and WRS testing.

As a consequence of the contradiction in results, the reliability of the said results must be discussed. A case can be formed against the RMSE as a least squares score, owing to its dependence on the assumption of Gaussianity and the conceptual link to the MSE model in that way. The inability of the CLF models to perform comparatively to the MSE model for even extreme values of $\sigma$ may reflect a bias in least squares scoring towards the way in which the MSE facilitates learning. In contrast, the MSE and CLF models could be shown to perform similarly for larger extents of  Gaussian noise in MAE scoring. This may speak to less bias towards the MSE model as a consequence of being based on different statistical assumptions.

The performance of the MSE model seemed to be almost completely invariant to changes in the value of $\sigma$ (refer to Table~\ref{table:HCTable_Control_Gaussian}). Its performance remaining constant in the absence and presence of all considered extents of Gaussian noise reflects a potentially unforeseen consequence of introducing deterministic noise into the dataset. The additive noise introduced may have been insufficient to alter the performance seen in light of the deterministic noise. Comparatively, the CLF models were more responsive to the extent of Gaussian noise present in the data. 

Nonetheless, a behaviour of the CLF models that was identified in both scores is that their performances collectively improved with $\sigma$ until a certain threshold, after which their performances degraded. This was seen in the MAE scores, where their best performances were reported for $\sigma=10$ (see Fig.~\ref{fig:HC8_Gaus_MAE}), and for the RMSE scores, where their best performances were reported for $\sigma=50$ (see Fig.~\ref{fig:HC8_Gaus_RMSE}).  This was likely owing to the limited extents of Gaussian noise having a 'smoothing' effect on the oscillations of the data. This ultimately made the data easier for the CLF models to learn, somewhat countering the effects of the deterministic noise.

The KW and WRS tests provided evidence ($p<0.05$ for all scores) of differing performances between the CLF models themselves for varying additive Gaussian noise. Moreover, the models that performed best before the general performance turning point were not necessarily the models that performed best afterwards. This speaks again to the importance of appropriate CLF constant selection, but also how the most appropriate CLF constant may change based on the noise present in the data. In this case, the appropriate CLF constant may be linked to the data smoothing taking place. 

\begin{figure}[!ht]
    \centering
    \caption{MAE scores for models varying with scale parameter $\tau$ of Cauchy noise (8-variable handcrafted experiment), and with varying proportion of simulated outliers in Seoul bike demand dataset.}
        \subfloat[8-variable handcrafted experiment]{\includegraphics[width=0.4\columnwidth]{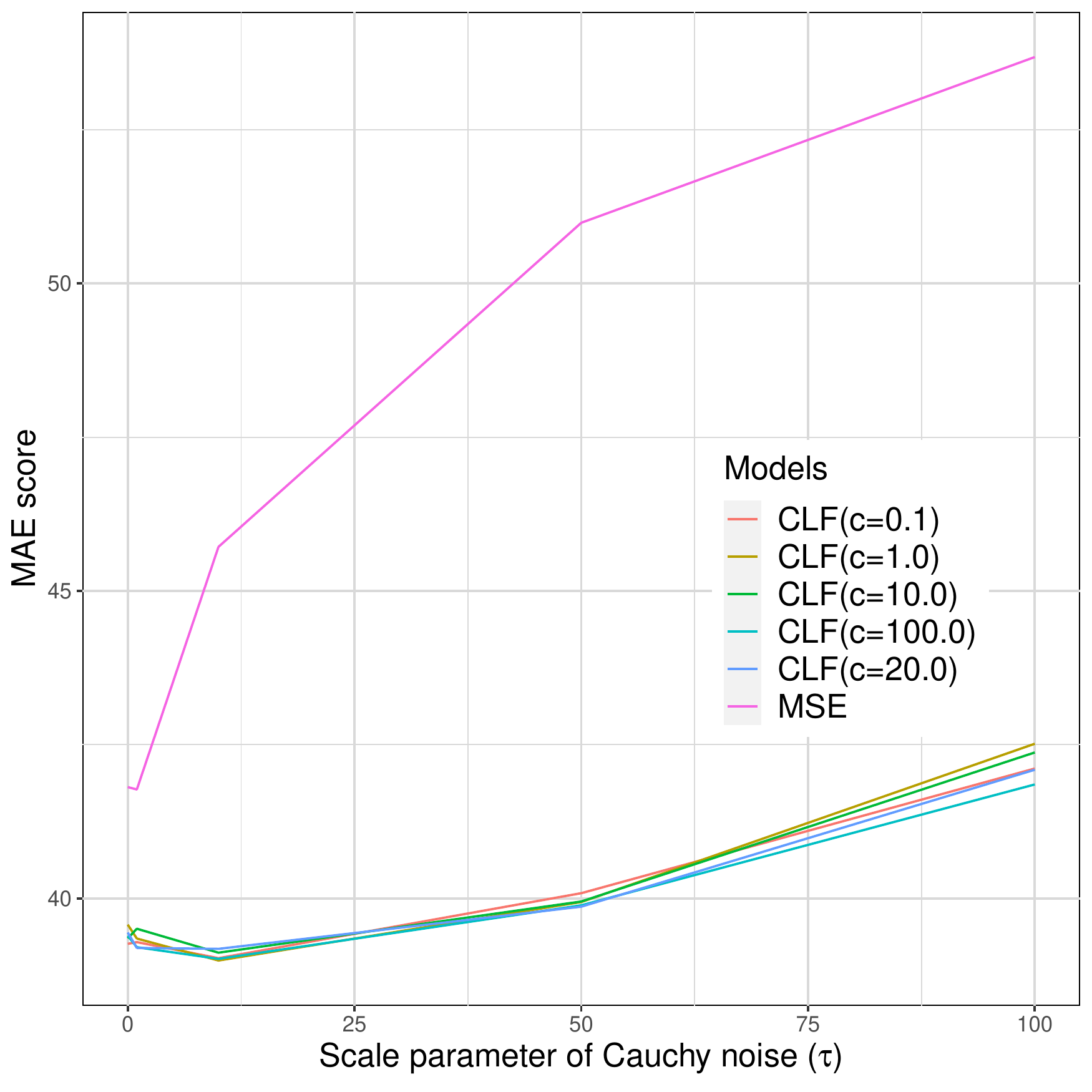}}\label{fig:HC8_Cauc_MAE}
        \qquad
        \subfloat[Seoul bike demand]{\includegraphics[width=0.4\columnwidth]{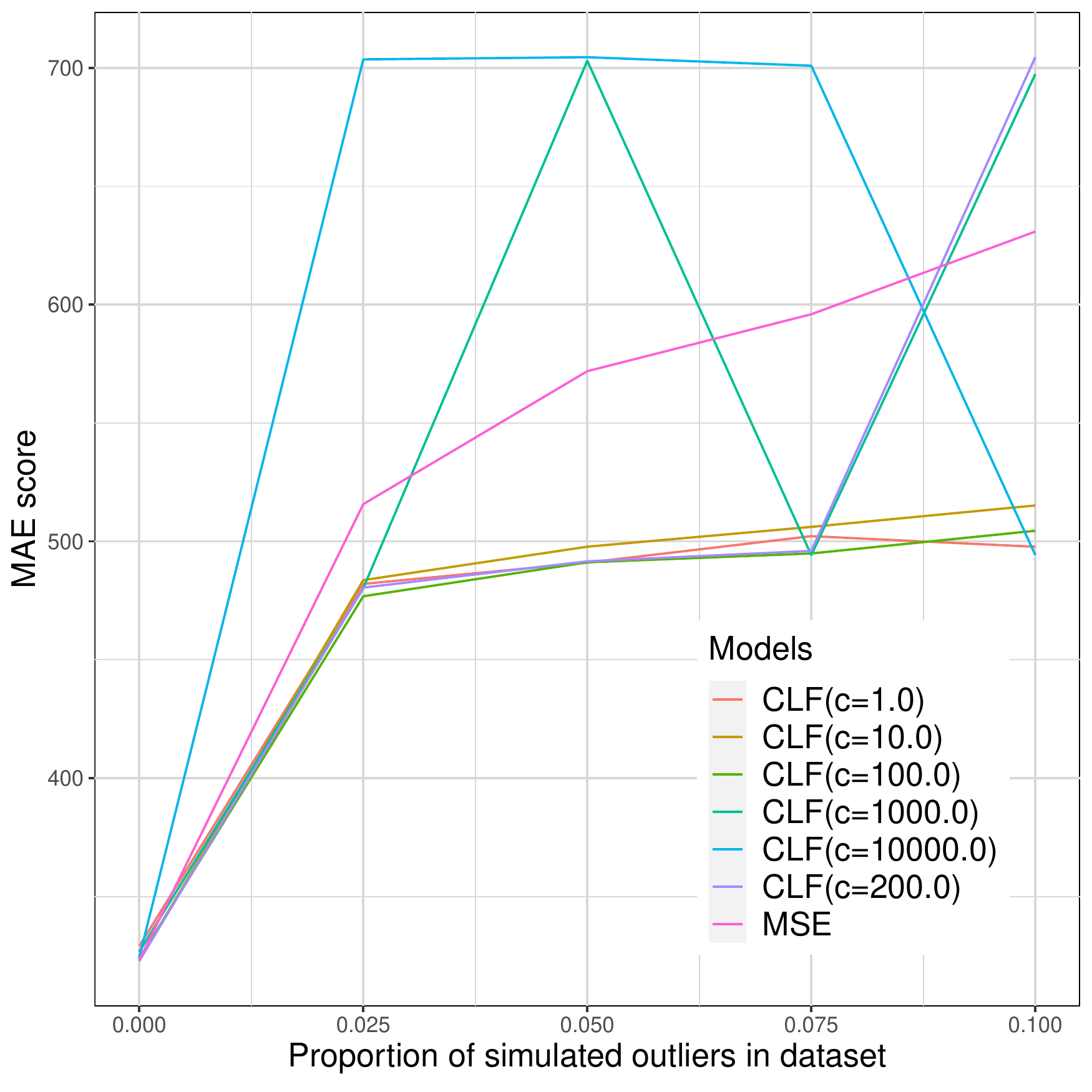}}\label{fig:bike_MAE}
    \label{fig:subfigname}
\end{figure}

The Cauchy configuration of the experiment behaved in a much more predictable fashion than the Gaussian one (see Table~\ref{table:HCTable_Cauchy}). It is worth noting once more that the MAE score will be the only one used as evidence for inference, but the RMSE scores align closely with them. For $\tau=10$ and above, the MSE model's performance degraded significantly in both scores (refer to Fig. \ref{fig:HC8_Cauc_MAE}). KW and WRS testing asserted that there was significant evidence ($p<0.005$ in MAE score) to indicate that the CLF models outperformed the MSE model for all values of $\tau$ above 10. 

Looking specifically at the CLF models, we see a similar behaviour to the Gaussian configuration; the performance of CLF models increased with $\tau$ until a certain threshold, after which their performance degraded, as shown in Fig.~\ref{fig:HC8_Cauc_MAE}. This behaviour was likely due to some degree of smoothing owing to the additive noise that made the oscillations in the data easier to learn. This ultimately counteracted the deterministic noise, until the additive noise itself began to dominate the signal. At this point, performance degraded; but unlike with the MSE model, the CLF models' performances degraded more gradually. 
There was suggestive evidence ($p<0.1$ in MAE) of differences between the CLF models at all values of $\tau$, according to WRS testing. This suggests that the combination of additive and deterministic noise makes it harder for a single CLF constant to be considered optimal. 

\subsection{Seoul bike sharing demand experiment}

Results obtained by corrupting various proportions of the training data with simulated outliers were largely in line with expectations (see Table~\ref{table:SeoulTable}). As a negative control, the initial configuration saw all seven models trained and tested on uncorrupted data, and neither KW nor WRS testing revealed significant evidence to suggest different performances between CLF and MSE for any of the scores. 

\begin{table*}[h] 
\centering
\scriptsize
\setlength{\tabcolsep}{0.3pt}
\caption{Seoul bike sharing demand experiment. Smallest error value for each setup is shown in red.}
\label{table:SeoulTable}
\begin{tabular}{|c|cc|cc|cc|}
\hline
        & \multicolumn{2}{|c}{Negative (0.000)} & \multicolumn{2}{c}{0.025} & \multicolumn{2}{c|}{0.05} \\
\hline
Model             & MAE         & RMSE        & MAE         & RMSE        & MAE        & RMSE        \\
             \hline
CLF 1.0      & 329.01 (12.76) & 469.19 (18.32) & 482.02 (5.07) & 618.75 (5.104) & \textbf{\textcolor{red}{491.03 (7.25)}} & 635.72 (1.20)  \\
CLF 10.0     & 322.85 (1.75) & \textbf{\textcolor{red}{459.64 (1.19)}} & 483.63 (9.11) & 621.61 (4.85) & 497.75 (9.91) & 635.74 (6.49)  \\
CLF 100.0    & 323.69 (1.82) & 460.42 (1.55) & \textbf{\textcolor{red}{476.8 (3.79)}} & \textbf{\textcolor{red}{614.59 (6.23)}}  & 491.14 (3.93) & 635.59 (5.65) \\
CLF 200.0    & \textbf{\textcolor{red}{322.75 (1.98)}} & 460.43 (0.81) & 480.47 (4.14) & 616.84 (1.43) & 491.57 (4.85) & \textbf{\textcolor{red}{630.43 (5.36)}}   \\
CLF 1,000.0  & 326.67 (2.19) & 462.21 (1.23) & 480.21 (5.24) & 615.29 (3.45) & 702.99 (2.97) & 953.56 (2.65) \\
CLF 10,000.0 & 324.25 (2.60) & 460.39 (1.75) & 703.57 (1.94) & 953.99 (1.74) & 704.49 (0.12) & 954.85 (0.07) \\
MSE          & 323.16 (1.71) & 461.45 (1.65) & 515.71 (15.61) & 674.29 (16.76) & 571.89 (16.64) & 751.85 (17.49)  \\
\hline
        & \multicolumn{2}{|c}{0.075} & \multicolumn{2}{c|}{0.100} &\\ 
\cline{1-5}
Model             & MAE         & RMSE        & MAE         & RMSE    &    \\
\cline{1-5}
CLF 1.0      &   502.19 (9.57) & 658.37 (9.12)  & 497.72 (5.33) & 651.63 (7.97)&  \\ 
CLF 10.0     &   506.11 (10.49) & 649.22 (9.31) & 515.16 (15.55) & 658.05 (14.93) &\\ 
CLF 100.0    &   494.88 (9.58) & 642.99 (8.36) & 504.47 (4.61) & 652.59 (2.88) &\\
CLF 200.0    &  495.97 (3.19) & 640.2 (4.95) & 704.55 (0.04) & 954.94 (0.06) & \\
CLF 1,000.0  &  \textbf{\textcolor{red}{494.16 (6.81)}} & \textbf{\textcolor{red}{639.24 (8.83)}}  & 697.36 (8.06) & 947.05 (9.26)  & \\
CLF 10,000.0 &    700.89 (5.02) & 951.44 (4.7)   & \textbf{\textcolor{red}{494.21 (5.75)}} & \textbf{\textcolor{red}{642.88 (5.42)}} & \\
MSE          &  595.93 (25.18) & 783.75 (22.75) & 630.86 (47.43) & 815.79 (57.71)  &   \\ 
\cline{1-5}
\end{tabular}
\end{table*}

However, as the proportion of outliers increased, the performance of the models began to shift. Firstly, the MSE model's performance degraded noticeably in comparison to the performance of the CLF models. From the 2.5\% proportion of noise onward, significant evidence ($p<0.001$ for all scores) was found using the WRS test suggesting that the MSE model was outperformed by the CLF models (refer to Fig. \ref{fig:bike_MAE}). 
The rate of performance degradation as outliers increased was higher for the MSE model than for the high-performing CLF models.

The models trained with the Cauchy loss function with reasonable CLF constants (less than or equal to 100) consistently performed better than the MSE model. This supports the hypothesis that CLF works well in the presence of outliers. It should be noted that the outliers were generated with a distribution that is very different from the Cauchy distribution. The models trained with extreme CLF constants (greater than or equal to 200) exhibited odd behaviour, sometimes performing better and sometimes performing worse than the MSE model. This highlights the sensitivity to the choice of CLF constants, hyper-parameters (such as the learning rate), and the optimiser.

\section{Conclusion}\label{sec:conclude}

Examining the underlying statistical assumptions in machine learning provides insight and possible improvements for practical applications. This study illustarted the shortcomings of the mean squared error (MSE) that arise owing to the implicit assumption of Gaussian noise. This assumption is prone to poor performance when the distribution of random noise in a dataset can be modelled by heavier-tailed profiles than the Gaussian, or when the data has a large number of outliers. The Cauchy loss function (CLF) was studied as an alternative, and was shown to be capable of performing well for its own assumed noise profile, as well as generalising better to the Gaussian noise profiles than the MSE could generalise to Cauchy noise profiles. CLF was also shown to be more robust to the combination of deterministic and additive noise in this study; promoting the learning of data in spite of, and even because of, the additive noise through smoothing. This is in contrast with MSE, which is heavily influenced by the deterministic noise. The CLF was also shown to perform better in the presence of outliers in a real dataset. 

The choice of the CLF constant significantly affects the performance of CLF. The means by which to choose an appropriate CLF constant, learning rate, optimiser and hyperparameter optimisation fall outside of the scope of this paper, and is suggested for future research.
Other recommendations for future research include a recreation of similar tests, but with more powerful inferential tools and more data.




\bibliographystyle{ieeetr}
\bibliography{myBibFile}

\end{document}